\pdfoutput=1
\documentclass[11pt]{article}

\usepackage[final]{acl}
\usepackage{times}
\usepackage{latexsym}

\usepackage[T1]{fontenc}
\usepackage[utf8]{inputenc}

\usepackage{microtype}

\usepackage{inconsolata}

\usepackage{graphicx}

\title{IntegrityAI at GenAI Detection Task 2: Detecting Machine-Generated Academic Essays in English and Arabic Using ELECTRA and Stylometry}

\author{Mohammad AL-Smadi \\
    Digital Learning and Online Education Office \\
  Qatar University, Doha, Qatar \\
  \texttt{malsmadi@qu.edu.qa} \\
}

\begin{document}

\maketitle
\begin{abstract}
Recent research has investigated the problem of detecting machine-generated essays for academic purposes. To address this challenge, this research utilizes pre-trained, transformer-based models fine-tuned on Arabic and English academic essays with stylometric features. Custom models based on ELECTRA for English and AraELECTRA for Arabic were trained and evaluated using a benchmark dataset. Proposed models achieved excellent results with an F1-score of 99.7\%, ranking 2\textsuperscript{nd} among of 26 teams in the English subtask, and 98.4\%, finishing 1\textsuperscript{st} out of 23 teams in the Arabic one. 
\end{abstract}

\section{Cite this work}

Mohammad AL-Smadi. 2025.
IntegrityAI at GenAI Detection Task 2: Detecting Machine-Generated Academic Essays in English and Arabic Using ELECTRA and Stylometry. In Proceedings
of the 1st Workshop on GenAI Content Detection
(GenAIDetect), Abu Dhabi, UAE. International Con-
ference on Computational Linguistics. In Proceedings
of the 1st Workshop on GenAI Content Detection
(GenAIDetect), Abu Dhabi, UAE. International Con-
ference on Computational Linguistics.

\section{Introduction}

Since the launch of ChatGPT in November 2022, research on developing models for artificial intelligence (AI)-generated text detection has increased. This increase reflects growing concerns about maintaining academic integrity in the face of advanced generative AI (GenAI) tools capable of producing human-like text \cite{al2023chatgpt}. \citet{48} were among the first working on this topic by developing a dataset named the "Human ChatGPT Comparison Corpus (HC3)" out of nearly 40,000 questions from different datasets along with their answers provided by humans. They also generated responses to these questions by ChatGPT and used the combined dataset to train detectors in both English and Chinese. The developed models included machine and deep learning based models like RoBERTa \citep{liu2019roberta} and demonstrated decent performance across different scenarios. 

Another paper focused on detecting ChatGPT-generated text written in English and French \citep{55}. The English model was trained using the HC3 dataset. The authors also translated some of its English content to French and included additional small French out-of-domain dataset of 113 French responses from ChatGPT and 116 from BingGPT. They fine-tuned two pre-trained models, CamemBERT \citep{57} and CamemBERTa \citep{58}, using the French dataset, and RoBERTa \citep{liu2019roberta} and ELECTRA \citep{clark2020electra} models using the English one. They also used XLM-R \citep{59} as multi-language model for the combined datasets of both languages. Research results showed that all models demonstrated good performance in identifying machine-generated content within the same domain, but when tested on out-of-domain content, their results dropped.

\citet{alshammari2024ai} used transformer-based models, namely AraELECTRA \citep{antoun2020araelectra} and XML-R \citep{59} to solve the challenges of machine-generated Arabic text identification. The authors focused on the influence of diacritics on detection model performance. Their method showed great accuracy on the AIRABIC benchmark dataset. Other research utilized stylometric features to detect machine-generated content. For instance, \citet{Kutbi2024DetectingCC} introduced a machine learning model with stylometry for identifying "Contract cheating", the act of students depending on others to complete academic assignments on their behalf, by detecting deviations from a learner's distinctive writing style, which achieved excellent accuracy in their research. \citet{10.1007/978-3-031-64312-5_13} developed a data-driven model named "StyloAI" trained with 31 stylometric features to detect machine-generated content. "StyloAI" performance outperformed other models on the same dataset.

%In 2023, research relied on watermarking techniques to facilitate machine-generated content detection \citep{51,52,53}. These research examples utilized watermarking based on embedding green tokens or cryptography techniques embedded in generated text from large language models (LLMs) as machine-detectable signals that are invisible to human readers.
\begin{table*}[ht]
\centering
\begin{tabular}{lllll}
\hline
\textbf{Language} & \textbf{Train Size} & \textbf{Dev. Size} & \textbf{Eval. Size} & \textbf{Test Size} \\
\hline
Arabic & 2070 (AI: 925, Human: 1145) & 481 (AI: 299, Human: 182) & 886 & 293 \\
English & 2096 (AI: 1467, Human: 629) & 1626 (AI: 391, Human: 1235) & 869 & 1130  \\
\hline
\end{tabular}
\caption{Summary of Arabic and English Datasets by subtask and type (Train, Development, Evaluation, and Test)}
\label{dataset}
\end{table*}
\citet{wee2023non} discovered that AI identification technologies classified human-written writings translated from non-English languages as AI-generated, which raised worries among non-native English speakers \citep{Liang2023GPTDA}. Moreover, \citet{weber2023testing} tested many AI text identification systems and found that they were neither accurate nor dependable, especially when content masking techniques were used. 

This research aims at addressing the challenge of AI-generated text. The rest of this paper is organized as follows: Section~\ref{method} discusses the research methodology, Section~\ref{results} presents the findings, and Section~\ref{conclusion} concludes the study and highlights future directions. %The rest of this paper is organized as follows, section~\ref{method} discusses the research methodology, by explaining the research task, dataset used, and developed models, section~\ref{results} discusses the research results and findings, and section~\ref{conclusion} concludes this research and sheds the light on future directions.

\section{Research Methodology}
\label{method}
\subsection{Task}
This research is based on our participation in the shared task "GenAI Content Detection Task 2: AI vs. Human – Academic Essay Authenticity Challenge" \citep{chowdhury2025genai}, which is organized as part of the "Workshop on Detecting AI Generated Content at the 31st International Conference on Computational Linguistics (COLING 2025)". The task aims at encouraging researchers to submit their research for detecting AI-generated academic essays. The task is designed to have three phases: (a) Models training and validation, (b) First evaluation phase, also referred as development phase, and (c) Models testing phase. Participated teams were ranked based on the results achieved in the final phase, i.e. models testing phase. The task covers content generated in two languages, Arabic and English. The next section explains in more detail the datasets provided for model training, validation, and testing.

\subsection{Dataset}
The datasets for this task consist of essays generated by generative AI models and human written ones. The essays authored by humans were curated from the "ETS Corpus of Non-Native Written English"\footnote{\url{https://catalog.ldc.upenn.edu/LDC2014T06}}, whereas the AI-generated ones were generated using seven different models including, GPT-3.5-Turbo, GPT-4o, GPT-4o-mini, Gemini-1.5, Llama-3.1 (8B), Phi-3.5-mini and Claude-3.5.

The datasets are designed to cover the three-phase task as discussed in the previous section. Each dataset consists of, (a) a train dataset for models training, (b) a validation dataset to fin-tune models parameters and evaluate model's performance during training phase, (c) an evaluation dataset, which is used to evaluate the model performance in a controlled environment before the final testing phase, and (d) a test dataset for the final testing phase, which is designed to assess the model's generalization and performance on completely unseen data.

Table~\ref{dataset} describes the sizes of the task's datasets by type and language. As described in the table, the Arabic dataset contains a balanced training set of 2,070 essays, which include 925 AI-generated essays, and 1,145 human-authored essays. This balance establishes an equitable foundation for training models to detect stylistic distinctions between AI and human text. In contrast, the English dataset has an imbalance in its training data, with 2,096 essays dominated by AI-generated texts (1,467 AI generated against 629 Human written). This skewed distribution may lead to model bias problem.
\begin{table*}[ht]
\centering
\begin{tabular}{llll}
\hline
\textbf{Feature} & \textbf{ELECTRA-Small} & \textbf{ELECTRA-Base} & \textbf{ELECTRA-Large} \\ \hline
Hidden Size & 256 & 768 & 1024 \\ 
Number of Layers & 12 & 12 & 24 \\ 
Number of Attention Heads & 4 & 12 & 16 \\ 
Total Parameters & 14 million & 110 million & 335 million \\ \hline
%\textbf{Memory Requirement} & Low & Moderate & High \\ \hline
%\textbf{Training Time} & Fast & Moderate & Slow \\ \hline
%\textbf{Inference Time} & Fast & Moderate & Slow \\ \hline
%\textbf{Performance} & Suitable for lightweight tasks, less complex patterns & Good performance on a wide range of NLP tasks & State-of-the-art performance, handles complex language patterns well \\ \hline
%\textbf{Use Case} & Mobile or edge devices, resource-constrained environments & General-purpose NLP tasks, suitable for moderate computational resources & High-stakes NLP tasks, research, or applications where top performance is needed \\ \hline
\end{tabular}
\caption{Comparison of ELECTRA-Small, ELECTRA-Base, and ELECTRA-Large models. AraELECTRA has the same features on ELECTRA-Base }
\label{tab:electra_comparison}
\end{table*}

\begin{table*}[ht]
\centering
\begin{tabular}{p{0.4\linewidth} p{0.5\linewidth}}
\hline
\textbf{Feature}                             & \textbf{Description}                                                                                   \\ \hline
Word Count                                   & Total number of words in the text.                                                                      \\ 
Sentence Count                               & Total number of sentences in the text.                                                                  \\ 
Average Sentence Length (words)              & Average number of words per sentence.                                                                   \\ 
Vocabulary Richness (Type-Token Ratio)       & Ratio of unique words to total words, indicating vocabulary diversity.                                   \\ 
Average Word Length (characters)             & Average number of characters per word.                                                                  \\ 
Commas                                       & Number of comma punctuation marks in the text.                                                          \\ 
Periods                                      & Number of period punctuation marks in the text.                                                         \\ \hline
\end{tabular}
\caption{Description of stylometric features extracted from the dataset.}
\label{tab:features}
\end{table*}

\subsection{Baseline Model}
The task organizers have implemented the following baseline model \citep{chowdhury2025genai}. For each language, a baseline model is trained using an $n$-gram approach, specifically unigrams. The textual content of the essays is transformed into a Term Frequency-Inverse Document Frequency (TF-IDF) representation, with the features limited to a maximum of 10,000. Finally, the performance is evaluated by training a Support Vector Machine (SVM) classifier on this feature representation.

\subsection{IntegrityAI Model}
The proposed model is based on ELECTRA \citep{clark2020electra} and its implementation named AraELECTRA \citep{antoun2020araelectra}, which is a model specifically tuned for the Arabic language. ELECTRA is an encoder only transformer that is designed to enhance the efficiency of implementing models for NLP tasks. Instead of implementing a masked language model (MLM), ELECTRA utilizes a unique training strategy known as "replaced token detection". While other encoder only transformers (such as BERT \citep{devlin2018bert}) implement MLM training strategy by predicting  masked words in a sentence, ELECTRA relies on its generator component to generate plausible alternatives to  replace some tokens in the input text. Then, uses the discriminator component to detect whether the token is replaced or original. The "replaced token detection" training strategy, requires the model to evaluate and learn all the input text tokens instead of the masked ones - as in BERT - which increases the model efficiency and minimizes the number of training epochs required to train the model. ELECTRA has three different pre-trained models that were used in this research, see Table~\ref{tab:electra_comparison} for  differences between them\footnote{\url{https://github.com/google-research/electra}}.  

\begin{figure}[t]
    \centering
  \includegraphics[width=1.0\linewidth]{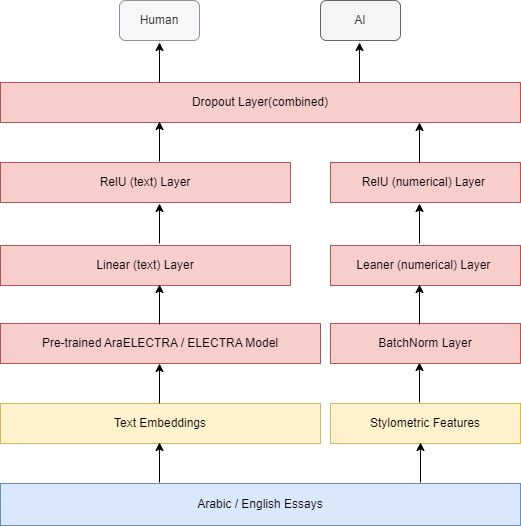} \hfill
  \caption {The  architecture of the ELECTRA-based models with stylometric features.}
  \label{model architecture}
\end{figure}

\begin{table*}[ht]
\centering
\begin{tabular}{lll}
\hline
\textbf{Model} & \textbf{Eval. Phase F1 (\%)} & \textbf{Testing Phase F1 (\%)} \\ \hline
AraELECTRA\_base\_discriminator & 99.8 & \textbf{98.4} \\ 
AraELECTRA\_base\_discriminator without features & - & 96.9 \\ 
Baseline-Arabic Model & 57.5 & 46.1 \\ \hline
ELECTRA\_small\_discriminator & 100.0 & 98.5 \\ 
ELECTRA\_small\_discriminator without features & - & 96.1 \\ 
ELECTRA\_large\_discriminator & 100.0 & \textbf{99.7} \\
Baseline-English Model & 29.8 & 47.8 \\ \hline
\end{tabular}
\caption{Evaluation (i.e. models' development phase) and testing results for Arabic and English developed models in comparison to the baseline model.}
\label{tab:model_performance}
\end{table*}

\begin{figure*}[t]
  \includegraphics[width=0.48\linewidth]{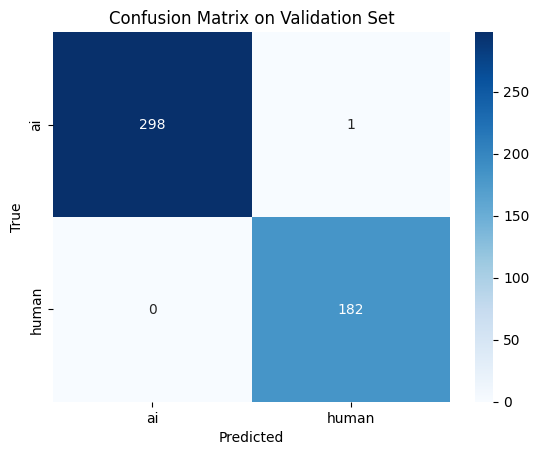} \hfill
  \includegraphics[width=0.48\linewidth]{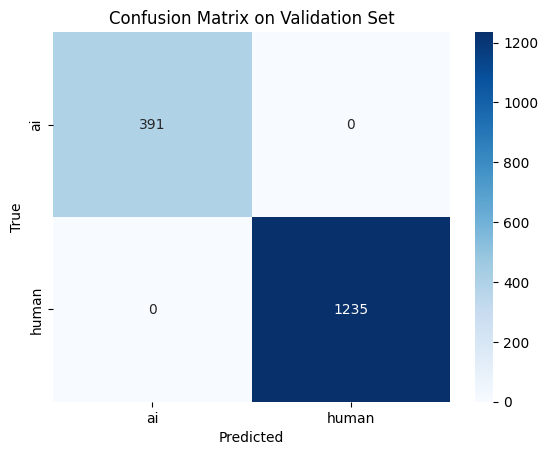}
  \caption {Confusion matrices on the validation sets (Arabic dataset on the left).}
  \label{confusionMatrix}
\end{figure*}
As depicted in Figure~\ref{model architecture}, the same model architecture was used for both the Arabic and English text classification. The ELECTRA model was trained on the English dataset, whereas its tuned version on Arabic, i.e., AraELECTRA was trained on the Arabic dataset. Both datasets went into a standard preprocessing phase, then stylometirc features were extracted and used with the text embeddings to train the pretrained models (see Table~\ref{tab:features} for more information about extracted features). The following layers were added to enhance the models' performance:

1. Dropout Layer: is a regularization technique where, during training, random neurons are temporarily ignored ("dropped out") to prevent overfitting and improve the model's generalization \cite{srivastava2014dropout}.

2. Batch Normalization ("BatchNorm1d"): normalizes the features of the input vector, stabilizing learning, and aiding in faster and more stable training \citep{ioffe2015batch}.

3. Fully Connected (Linear) Layers: these layers are basic neural network layers where every input is connected to every output by a learned weight. These layers include: (a) "numerical": takes the batch-normalized numerical features and projects them onto a new space to learn a higher-level representation of these features. (b) "text": processes the [CLS] token embedding from the ELECTRA model output, allowing the model to further tailor this representation for the task at hand. 

4. Rectified Linear Unit (ReLU) activation Function: is a non-linear operation used after linear layers to introduce non-linear properties to the model, making it capable of learning more complex patterns \citep{glorot2011deep}. This layer is used after each of the fully connected layers (numerical and text) to add non-linearity to the model, which helps in learning complex patterns in the data.

5. Output Layer (Fully Connected (Linear)): after processing through their respective pathways, both text and numerical data features are combined (concatenated) to form a unified feature vector. This combined feature vector is then passed to a final fully connected layer (combined), which outputs the logits for the classification categories. %These logits are then used to compute the loss during training and can be converted to probabilities for making predictions.

The models were trained for 10 epochs with the option of ( early\_stopping\_patience=2) implemented to avoid model overfitting during training. Models participating in this task were evaluated and ranked based on their achieved F1-score.

\begin{figure*}[t]
  \includegraphics[width=0.5\linewidth]{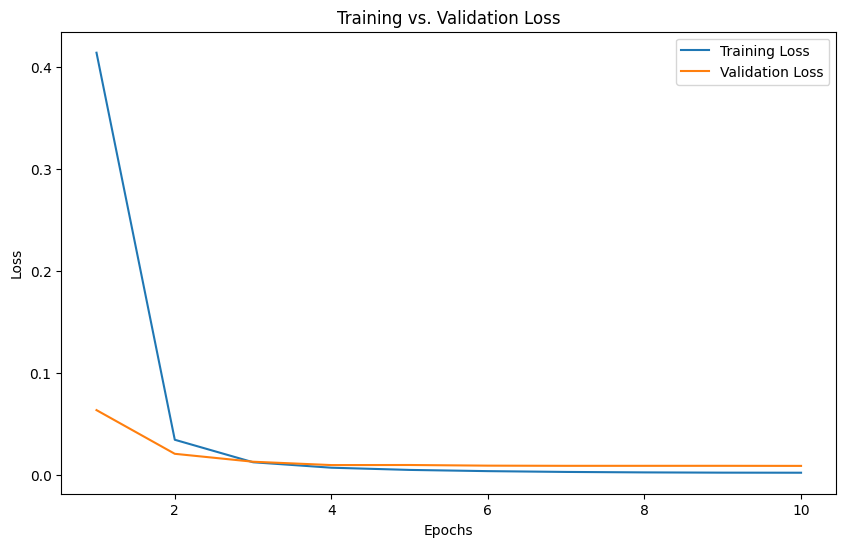} \hfill
  \includegraphics[width=0.5\linewidth]{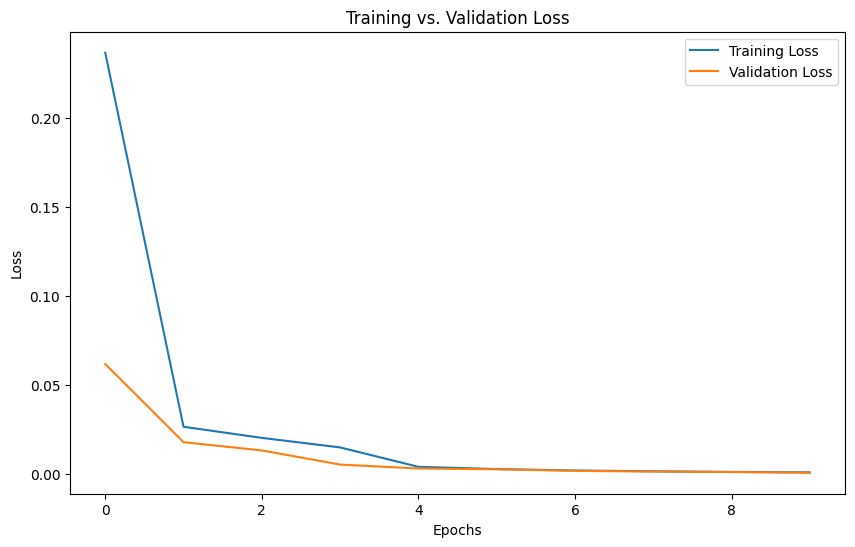} \hfill
  \includegraphics[width=0.6\linewidth]{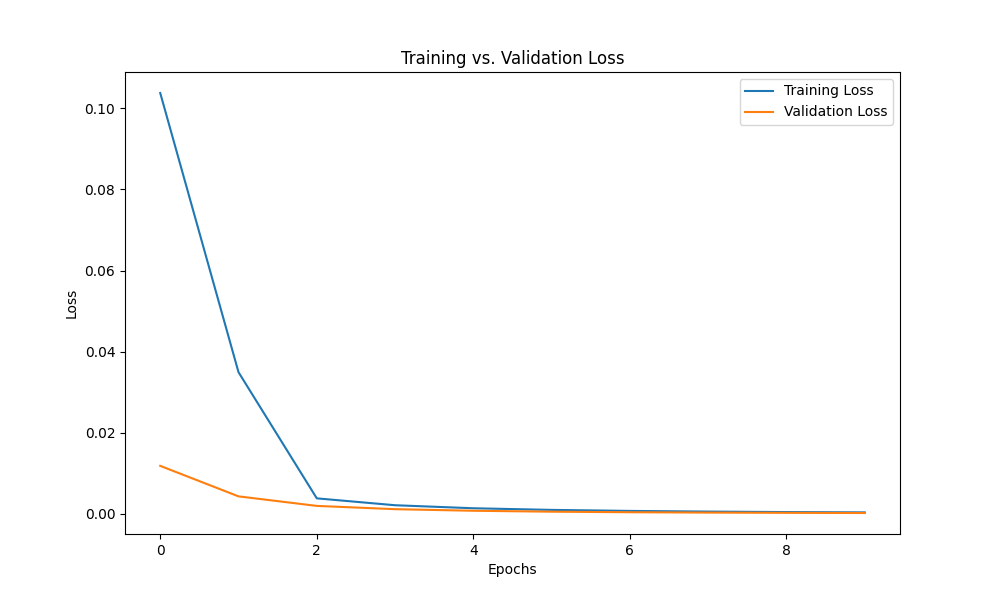} \hfill
  \caption {Training vs. validation loss values after each epoch of models training (AraELECTRA the upper left corner, ELECTRA\_small on the right upper corner, and ELECTRA\_large on the left lower corner)}
  \label{lossValues}
\end{figure*}
\section{Results and Findings}
\label{results}
Table~\ref{tab:model_performance} presents the developed models results for Arabic and English datasets. Results show that models achieved high F1 scores of 99.8\% for the Arabic dataset and 100\% for the English dataset in the evaluation phase and maintained that high performance in the testing phase with (98.4\% and 98.5\%, for Arabic and English datasets respectively). This achievement demonstrates that the models are not only well-tuned to the training data but also maintain their discriminative power on new and unseen data. This finding is also represented by the confusion matrices on the validation datasets. The trained model on the English dataset classified all 'ai' and 'human' labels accurately. Whereas, The trained model on the Arabic dataset had a near-perfect classification with only one instance of 'ai' being misclassified as 'human' (see Figure~\ref{confusionMatrix}).   

To evaluate the impact of the stylometric features on the model performance, we trained the models without features. The results demonstrate that excluding these features leads to a decrease in model performance, with a 1.5\% and 2.4\% drop in F1 score for AraELECTRA and ELECTRA models, respectively. This indicates that extracted features enhanced model predictions. Despite the modest decline, the impact underscores the importance of these features for better generalization.

The results of training vs. validation Loss values after each epoch of models training in Figure~\ref{lossValues}, show that the training loss rapidly declined from the first epoch and then quickly stabilized to run in parallel with the validation loss. Both values of training and validation loss kept decreasing smoothly until the end of models training epoch without any sign of overfitting, as the validation loss remains close to the training loss throughout the training process. This was also maintained by enabling the option of  early\_stopping during the models training. Moreover, this also indicates that both models generalizes very well when confronted by new unseen data. 

The rapid stabilization of loss values may indicate that more complex model architectures might achieve even better results. Therefore, we trained the ELECTRA\_large instead of the ELECTRA\_small model for the english subtask for 10 epochs as well. As, expected the ELECTRA\_large achieved better results with F1 score of 99.7\%.

For more information on  the results of other participating teams in the task, the reader is redirected to \citep{chowdhury2025genai}.

\section{Conclusion and Future Work}
\label{conclusion}
This study demonstrates the efficacy of transformer-based models for identifying machine-generated academic articles. Using ELECTRA-Small for English and AraELECTRA-Base for Arabic, paired with stylometric characteristics, our models produced remarkable F1-scores of 98.5\% and 98.4\%, respectively. Experiments using ELECTRA-Large for English revealed the possibility of even better F1-score, reaching 99.7\%, but at a larger computing cost.

Our proposed models offer an adaptable solution that balances performance and efficiency and is appropriate for a variety of hardware setups. To improve robustness, future study might focus on real-time detection, expanding to new academic areas, and extending language coverage.

%\section*{Acknowledgments}

\bibliography{custom}

\end{document}